\documentclass[11pt,a4paper]{article}
\usepackage[T1]{fontenc}
\usepackage[utf8]{inputenc}
\usepackage[english]{babel}
\usepackage[a4paper,margin=25mm]{geometry}
\usepackage{amsmath,amssymb}
\usepackage{newtxtext,newtxmath}
\usepackage{microtype}
\usepackage{graphicx}
\usepackage{float}
\usepackage{placeins}
\usepackage{booktabs}
\usepackage{enumitem}
\usepackage[font=small,labelfont=bf]{caption}
\usepackage[numbers,sort&compress]{natbib}
\usepackage{xurl}
\usepackage[hidelinks]{hyperref}
\graphicspath{{figures/}}
\hypersetup{
  pdftitle={Meta-Initialization of Implicit Neural Representations to Accelerate Deformable Medical Image Registration},
  pdfauthor={Marat O. Zinovev and Dmitry V. Sorokin},
  pdfsubject={Deformable medical image registration and meta-learning},
  pdfkeywords={implicit neural representations, meta-initialization, Reptile, STRAINER, KAN-IDIR}
}
\setlist[itemize]{topsep=4pt,itemsep=2pt,parsep=0pt}
\newcommand{\MetricPlot}[2]{%
  \includegraphics[width=0.82\linewidth]{#1}\\[-2pt]
  {\small Adaptation iterations}\\[3pt]
  {\small #2}\par}
\newcommand{\DicePlot}[1]{\MetricPlot{#1}{(a) Dice}\medskip}
\newcommand{\HDPlot}[1]{\MetricPlot{#1}{(b) HD95}}

\title{\LARGE\bfseries Meta-Initialization of Implicit Neural Representations\newline
  to Accelerate Deformable Medical Image Registration\thanks{This is an
  English-language author version of a paper presented at GraphiCon 2026,
  Saint Petersburg, Russia, 22--24 September 2026. The Russian-language
  proceedings are forthcoming. The related-work discussion has been updated
  and several methodological descriptions have been clarified for this
  version.}}
\author{%
  Marat O. Zinovev\qquad Dmitry V. Sorokin\thanks{Corresponding author: \href{mailto:dsorokin@cs.msu.ru}{dsorokin@cs.msu.ru}.}\\[5pt]
  {\small Faculty of Computational Mathematics and Cybernetics}\\
  {\small Lomonosov Moscow State University, Moscow, Russia}\\[3pt]
  {\small\href{mailto:zinovevmo@my.msu.ru}{zinovevmo@my.msu.ru}\quad
  \href{mailto:dsorokin@cs.msu.ru}{dsorokin@cs.msu.ru}}}
\date{}

\begin{document}
\maketitle
\begin{abstract}
Implicit neural representation (INR) methods for deformable medical image
registration require the network parameters to be optimized for each new
image pair, making adaptation speed a practical limitation. This work
investigates whether early adaptation can be accelerated using an
initialization learned from a collection of previously solved registration
tasks. We examine two schemes: Reptile meta-initialization and STRAINER
transfer of the shared component of a coordinate network. KAN-IDIR is used
as the base INR displacement-field model. Experiments are conducted on
OASIS-1, with 364 consecutive pairs used for training and 49 pairs for
test-time adaptation. Registration quality is evaluated using mean Dice
and HD95, together with the fraction of voxels with a negative Jacobian
determinant at iteration 1500. In the configurations studied, both schemes
yield higher Dice and lower HD95 at early iterations than random
initialization. Reptile numerically outperforms STRAINER under a small
iteration budget, although their pretraining budgets are not matched.
As the number of iterations increases, the differences between the schemes
shrink. STRAINER pretraining takes 39--40 minutes, compared with 44.98 hours
for Reptile.
\end{abstract}

\section{Introduction}

Deformable image registration (DIR) is a fundamental problem in medical image
processing. Its objective is to determine a spatial transformation that aligns
one image with another so that corresponding anatomical regions occupy the
same spatial locations. This enables correct comparison, fusion, and
subsequent analysis of images in clinical practice.

Before the advent of deep learning, deformable registration relied primarily
on iterative optimization methods: Free-Form Deformation (FFD)
\cite{Rueckert1999}, LDDMM \cite{Beg2005}, Demons \cite{Thirion1998}, Elastix
\cite{Klein2009}, ANTs \cite{Avants2011}, pTV \cite{Vishnevskiy2016}, and Flash
\cite{Zhang2019}. These methods are well studied and mathematically grounded,
but they require parameter tuning for each image pair and incur substantial
computational cost. With the development of deep learning, direct
image-to-image architectures based on convolutional neural networks (CNNs) and
transformers have become increasingly common
\cite{Balakrishnan2019a,Chen2022}. Given an image pair, these methods directly
predict a displacement vector field and operate several orders of magnitude
faster than classical iterative methods. However, their performance depends
strongly on large training datasets, which are not always available for a
particular organ or modality, and the models themselves are sensitive to data
distribution shift.

An alternative approach models the displacement field as a mapping from
continuous voxel coordinates to the corresponding displacement vectors. This
mapping is parameterized by a neural network and treated as an implicit neural
representation (INR). INR methods have demonstrated outstanding performance
in novel-view synthesis (NeRF) \cite{Mildenhall2020} and have been successfully
applied in several DIR methods \cite{Wolterink2022,Harten2023,Harten2024}.
INR-based solutions mitigate key limitations of trained models: they reduce
dependence on large training sets, are less sensitive to data distribution
shift, and are more robust to differences in spatial resolution. Their main
drawback, however, is that the coordinate-network parameters must be optimized
anew for every image pair, typically from a random initialization.
Computational efficiency and optimization stability therefore become
critical, and adaptation speed for a new pair is a practical bottleneck.

Meta-learning offers one possible way to accelerate this adaptation. Instead
of a random starting point, one may use an initialization learned from a set of
previously solved registration tasks in the expectation that it will converge
faster on a new pair. This work investigates how much such an approach
accelerates adaptation in medical DIR and how it affects the final
registration quality. The main contributions are as follows:
\begin{itemize}
  \item we study meta-initialization of a KAN-based INR displacement-field
  representation for deformable medical image registration;
  \item we examine and compare two meta-initialization schemes: Reptile
  \cite{Nichol2018} and STRAINER transfer of the shared coordinate-network
  component \cite{Vyas2024Strainer};
  \item experiments on OASIS-1 show that, in the configurations studied,
  both schemes yield higher mean Dice and lower mean HD95 during the early
  stage of INR adaptation, and we report the pretraining cost of each
  scheme.
\end{itemize}

\section{Related Work}

\subsection{Deformable Medical Image Registration}

Classical iterative registration methods are based on well-founded
mathematical approaches: B-spline free-form deformations
\cite{Rueckert1999}, geodesic flows of diffeomorphisms in LDDMM
\cite{Beg2005}, the diffusion-based Demons model \cite{Thirion1998},
isotropic total-variation regularization of displacement fields in pTV
\cite{Vishnevskiy2016}, and the Fourier approximation of Lie algebras in Flash
\cite{Zhang2019}. Although pair-specific optimization can achieve high
registration quality, these methods are computationally expensive and require
careful hyperparameter tuning.

Progress in deep learning has led to direct image-to-image methods based
primarily on CNNs. VoxelMorph \cite{Balakrishnan2019a} enabled fast
registration with CNNs trained without supervision to predict displacement
fields. Subsequent developments included probabilistic diffeomorphic modeling
\cite{Dalca2019}, pyramidal architectures \cite{Mok2020b}, topology-preserving
schemes \cite{Mok2020a,Greer2021,Tian2023}, and transformer- and MLP-based
architectures \cite{Chen2022,Meng2024}. Even when trained on large datasets,
however, such models remain vulnerable to data distribution shift
\cite{Jena2024,Tian2024} and do not naturally transfer to data with a
different spatial resolution without retraining \cite{Drozdov2024}.

\subsection{INR-Based Image Registration}

INR approaches model the displacement field $U$ as a continuous neural
function over a spatial domain $\Omega$. Because the network parameters are
optimized anew for each image pair, these methods are computationally more
expensive than a single forward pass through a pretrained CNN. Nevertheless,
they offer several important advantages: no pretraining on large datasets,
lower sensitivity to data distribution shift, and resolution independence.
Implicit neural representations were first applied to medical image
registration in \cite{Wolterink2022}, where SIREN networks were used for lung
CT registration. The idea was subsequently extended to brain MRI
\cite{Byra2023} and developed into robust bidirectional schemes with
uncertainty estimation \cite{Harten2023}. The related NODEO approach
\cite{Li2021} formulates registration through neural ordinary differential
equations, whereas SINR \cite{SideriLampretsa2024} combines B-spline FFDs with
implicit neural representations. A common bottleneck of these methods is the
need to retrain the network for every pair, which makes computational
efficiency and optimization stability central concerns.

\subsection{Meta-Learning for INR}

A major limitation of INR is that the coordinate-network parameters are
usually optimized from a random initialization for every new signal or image
pair. Meta-learning treats this issue as learning over a distribution of
tasks: each task corresponds to a separate signal, image, or image pair, and
the goal is to find initial parameters or an update rule that enables rapid
adaptation of the INR to a new task with only a few gradient steps. Classical
methods of this kind include MAML \cite{Finn2017}, Meta-SGD
\cite{Li2017MetaSGD}, and the first-order Reptile algorithm
\cite{Nichol2018}. They are particularly natural for INR because test-time
adaptation is already formulated as optimization of a single coordinate
network.

Early studies showed that meta-learning can define a useful functional prior
for INR. MetaSDF \cite{MetaSDF2020} formulated signed distance function
reconstruction as learning a space of implicit representations. Tancik et
al.~\cite{Tancik2021} extended this idea to coordinate representations for
images, CT reconstruction, and NeRF scenes: meta-initialization accelerated
convergence and acted as a regularizing prior. Subsequent work considered
sparse initializations \cite{Lee2021MetaSparseINR}, INR-based image compression
\cite{Strumpler2022}, and the STRAINER framework for learning transferable
features, where shared initial INR layers are trained over a collection of
signals and transferred to a new coordinate network
\cite{Vyas2024Strainer}.

Meta-learning is especially relevant to deformable registration because
INR-based registration requires optimization for every new image pair. REINDIR
\cite{Harten2024Reindir} combines a template INR representation with an image
encoder that modulates hidden layers for a specific pair and provides a better
starting point for subsequent optimization on a test pair. cMeta-INR
\cite{Qian2026} extends this idea to registration from undersampled k-space
data using a population-level deformation prior.

Concurrent work, LUMINR \cite{rynczak2026luminr}, combines Reptile
meta-initialization with a SIREN whose inputs include spatial coordinates
and local image intensities and gradients. Its focus is the interaction
between local conditioning and meta-initialization. Our study keeps the
coordinate-only KAN parameterization fixed and compares Reptile with
STRAINER-based initialization, including their pretraining costs. The
studies therefore address complementary aspects of accelerating
pair-specific INR registration.

\section{INR Displacement-Field Model}

\subsection{INR Representation of the Displacement Field}

Deformable registration can be formulated as finding a deformation mapping
$\Phi(x)=x+U(x)$, where $U(x)$ is the displacement field. The mapping
minimizes the objective
\begin{equation}
  \hat{\Phi} = \arg\min_{\Phi}\;
  \mathcal{L}_{\text{data}}\bigl(M \circ \Phi,\, F\bigr)
  + \mathcal{L}_{\text{reg}}(U, \Phi),
  \label{eq:objective}
\end{equation}
where $\mathcal{L}_{\text{data}}$ measures similarity between the fixed image
$F$ and the warped moving image $M\circ\Phi$, and
$\mathcal{L}_{\text{reg}}$ regularizes the displacement field $U$ and the
mapping $\Phi$. Both images are assumed to be defined over the same domain,
$M,F\colon\Omega\subset[-1,1]^n\to\mathbb{R}$. In the INR approach, the
displacement field is parameterized by a coordinate network $U_\theta(x)$
that maps a voxel coordinate to a displacement vector. The parameters
$\theta$ are obtained by solving~\eqref{eq:objective} for a particular image
pair.

\subsection{Base Coordinate Model}

We use the KAN-IDIR coordinate network \cite{make8070184} as the base INR
displacement-field model. It is built on Kolmogorov--Arnold Networks (KANs)
\cite{Liu2024c} with Chebyshev polynomials as basis functions
\cite{Sidharth2024}. A Chebyshev polynomial of order $n$ is defined as
$T_n(x)=\cos\bigl(n\cdot\arccos(x)\bigr)$ for $x\in[-1,1]$, allowing the basis
to be evaluated efficiently in vectorized form. Before the polynomials are
evaluated, the input is normalized to $[-1,1]$ with a hyperbolic tangent, and
a trainable residual connection is added to each layer to stabilize
optimization. The network architecture is $[3,70,70,3]$, with two hidden
layers of 70 neurons, and a maximum polynomial degree of $D=28$, giving
159\,600 trainable parameters. The final deformation mapping is
$\Phi_\theta(x)=x+U_\theta(x)$. This coordinate network is used as a fixed
base parameterization of the displacement field. An additional comparison
uses RandKAN-IDIR~\cite{make8070184}: in each layer, the degree-zero polynomial
is supplemented with $k=12$ randomly selected degrees from 1 to $K=84$;
this architecture contains 74\,480 parameters. These models are used to
investigate how different parameter initialization schemes affect adaptation
speed on a new image pair.

\subsection{Loss Function}

\begin{figure}[htbp]
  \centering
  \includegraphics[width=\linewidth]{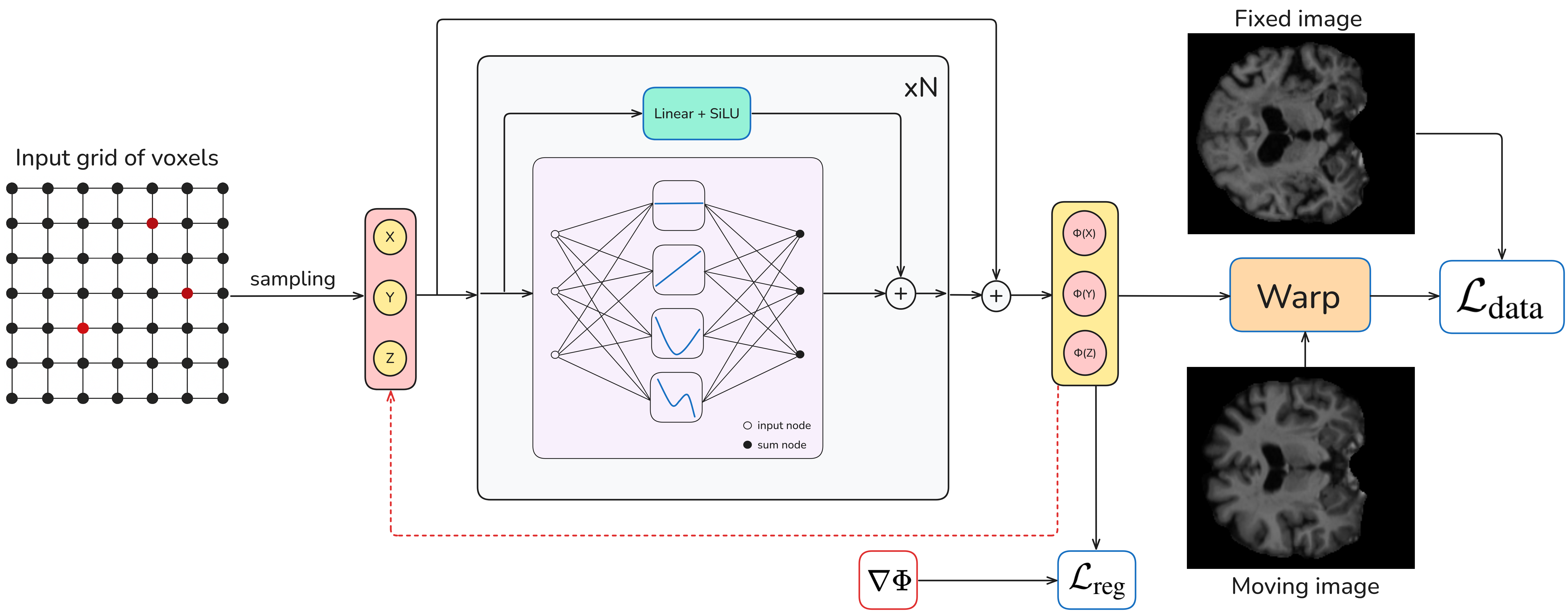}
  \caption{Overview of INR displacement-field optimization. Voxels are sampled
  from the input volume and passed to the coordinate network. The network
  predicts a displacement field that warps the moving image and enables
  computation of the data-similarity loss. For regularization, the
  displacement-field gradient is obtained by automatic differentiation.}
  \label{fig:pipeline}
\end{figure}

The loss is the sum of an image-similarity term and regularization terms
(see~\eqref{eq:objective}). Negative normalized cross-correlation,
$\mathcal{L}_{\text{data}}(I,J)=-\mathrm{NCC}(I,J)$, is evaluated over a
subset of voxels within the region-of-interest mask. Regularization comprises
a smoothness term based on total variation \cite{Rudin1992}, which penalizes
the $L_1$ norm of the displacement-field gradient, and a folding penalty
$\mathcal{L}_{J\!\det}$ for voxels with a negative Jacobian determinant
\cite{Mok2020a}. The complete loss is
\begin{equation}
  \mathcal{L} =
  \mathcal{L}_{\text{data}}
  + \lambda\,\mathcal{L}_{\text{smooth}}
  + \gamma\,\mathcal{L}_{J\!\det},
  \label{eq:loss}
\end{equation}
where $\lambda$ and $\gamma$ are weighting coefficients. The complete
displacement-field optimization pipeline is shown in
Fig.~\ref{fig:pipeline}.

\section{Meta-Initialization of the INR Displacement-Field Model}

\subsection{Random Initialization (Baseline)}

The baseline is the random initialization conventionally used for INR-based
registration. For each new image pair, all coordinate-network parameters
$\theta$ are randomly initialized and then optimized from scratch by
minimizing~\eqref{eq:loss}. The benefit of meta-initialization is evaluated
relative to this scheme.

Because the baseline network parameters are fitted independently for every
new pair, we examine two schemes for transferring information between
previously solved registration tasks in order to reduce the number of
adaptation steps and sensitivity to random initialization. In both cases,
meta-learning does not replace pair-specific optimization: the learned
parameters are used only as the initial state for subsequent fine-tuning. This
preserves the principal distinction between instance-optimized INR methods and
trained predictive models. The displacement field is not predicted directly
from the input images but remains the solution of a separate optimization
problem. Each meta-task $T_j$ registers a moving image $M^{(j)}$ to a fixed
image $F^{(j)}$ using the same loss $\mathcal{L}_{T_j}$ as the baseline. The
goal of meta-learning is to find an initial parameter point $\theta$ from
which optimization on a new pair converges faster and more reliably.

\subsection{Reptile Meta-Initialization}

Let $\theta$ denote the current shared initialization. A short inner
adaptation is performed from this point for each task $T_j$:
\[
  \theta_j^{(0)} = \theta,\qquad
  \theta_j^{(t+1)} =
  \theta_j^{(t)} - \alpha
  \nabla \mathcal{L}_{T_j}\!\left(\theta_j^{(t)}\right),
  \quad t=0,\dots,m-1,
\]
where $m=20$ inner steps are used in the experiments. For clarity, this
inner update is written schematically as plain gradient descent with step
size $\alpha$. In the experiments, each inner step is an Adam update
(Section~\ref{sec:protocol}). After adaptation, the
parameters are averaged across the set of meta-tasks,
$\bar{\theta}=\tfrac{1}{N}\sum_{j=1}^{N}\theta_j^{(m)}$, and the meta-update
moves the initialization toward the adapted parameters:
\[
  \theta \leftarrow \theta + \beta(\bar{\theta} - \theta).
\]
This first-order scheme \cite{Nichol2018} requires no second derivatives and
is therefore well suited to optimizing INR displacement-field
representations. Unlike conventional pretraining of one shared model, Reptile
uses the result of several inner optimization steps on each task and favors
initial parameters for which gradients from different batches of the same
pair produce consistent adaptation directions. In registration, the objective
is therefore not a universal displacement field but an INR parameter point
from which the field adapts rapidly to a particular pair.

\subsection{Transfer of the Shared INR Component (STRAINER)}

An alternative is to pretrain a shared encoder jointly with pair-specific
decoders and then transfer only the encoder to a new pair
\cite{Vyas2024Strainer}. Let
the INR be decomposed into a shared encoder $f_\theta$ and a pair-specific
decoder $g_\phi$:
$h_{\theta,\phi}(\mathbf{p})=g_\phi\!\bigl(f_\theta(\mathbf{p})\bigr)$. For the
$[3,70,70,3]$ architecture, the shared component contains every layer except
the last, accounting for 153\,300 of 159\,600 parameters (approximately
96.1\,\%). The output layer remains specific to an image pair. During
pretraining, the same encoder is used for a set of training pairs and each
pair has its own decoder:
\[
  \Theta^* =
  \arg\min_{\Theta}\;
  \frac{1}{|\mathcal{S}_t|}\sum_{i\in\mathcal{S}_t}
  \left(
    \mathcal{L}_{\mathrm{data}}\!\bigl(M^{(i)}\circ\Phi_i,\;F^{(i)}\bigr)
    + \mathcal{L}_{\mathrm{reg}}(U_i,\Phi_i)
  \right),
  \quad
  \Theta=[\theta,\,\{\phi_i\}_{i\in\mathcal{S}_t}].
\]
After pretraining, only $\theta^*$ is retained. For a new pair, the encoder is
initialized with these weights, a new decoder is created, and the entire
model is fine-tuned for the particular task. Separate decoders prevent the
shared component from simply memorizing one displacement field and force it
to extract features useful for rapid subsequent adaptation. The
shared component is not frozen during adaptation: the retained weights
$\theta^*$ provide only the initial parameterization, after which all INR
parameters are optimized together with the new decoder. This prior is most
useful when the distribution of new pairs is close to the training-data
distribution; transferability may decrease under a strong data distribution
shift.

Neither scheme changes the definition of the displacement field or the loss.
They are additional mechanisms for accelerating the early phase of INR
optimization while retaining the key property of the approach: the final
model is still adapted to a particular image pair.

\section{Experiments and Results}

\subsection{Data and Evaluation Metrics}

Experiments were conducted on OASIS-1 \cite{Marcus2007}, which contains
414 affinely aligned brain MRI volumes, one per subject, with labels for
35 major anatomical regions. The data were preprocessed and resized to
$160\times192\times224$ following \cite{Hoopes2022}.

For each pair, the moving-image segmentation is warped using nearest-neighbor
interpolation. Dice is computed separately for each non-background label
present in both segmentations; empty structures are excluded, and paired
bilateral labels are not merged. Scores are averaged first over structures
within each pair and then over the 49 test pairs. The region of interest for
loss computation is the union of all nonempty (non-background)
segmentations. Mean 95th-percentile Hausdorff distance (HD95) is additionally
evaluated at the same checkpoints, and the fraction of voxels with a negative
Jacobian determinant is evaluated at iteration 1500. The standard deviation
of the log-Jacobian determinant (SDlogJ) and deformation-field regularity at
early iterations were not measured.

The implementation uses PyTorch \cite{Paszke2019}. Training employs Adam
\cite{Kingma2014} with an initial learning rate of $10^{-4}$. At every
iteration, 10\,000 points are randomly sampled from the region of interest.
The loss weights are fixed at $\lambda=0.4$ and $\gamma=15$. All experiments
were performed on a server equipped with an NVIDIA A6000 GPU.

\subsection{Meta-Training and Adaptation Protocol}
\label{sec:protocol}

Evaluation uses a split of OASIS-1 into consecutive image pairs. The training
set contains 364 pairs, $(0,1),(1,2),\ldots,(363,364)$, and is used for
Reptile meta-training and STRAINER pretraining. The test set contains 49
pairs, $(364,365),(365,366),\ldots,(412,413)$. At test time, each of the
three schemes (random initialization, Reptile, and STRAINER) defines only
the starting point. The INR is then optimized for each test pair with the
same hyperparameters, and quality is measured as a function of the number
of adaptation iterations.

Reptile meta-training used 300 outer iterations with $\beta=0.2$. The
meta-batch comprised 364 tasks corresponding to the consecutive image pairs
$(0,1),\ldots,(363,364)$. Each task used 20 inner steps, for a total of
$300\times364\times20=2\,184\,000$ inner-optimization steps. The inner
optimizer was Adam with a learning rate of $10^{-4}$ and the default
PyTorch parameters. No separate outer optimizer was used: the parameters
were explicitly updated according to the Reptile rule with $\beta=0.2$
and no momentum. Complete meta-training took approximately 2699 minutes,
or 44.98 hours (44 hours 59 minutes).

STRAINER pretraining consisted of 5000 Adam steps with a learning rate of
$10^{-4}$, the default PyTorch parameters, and a constant learning rate.
At each step, 10 image pairs were randomly sampled from the training set
to form a task batch. For each pair, 10\,000 coordinates were sampled,
giving a total of up to 100\,000 coordinates per step. Pretraining took
approximately 39--40 minutes (about 0.66 hours).

Thus, the comparison concerns specific configurations with unmatched
pretraining budgets: Reptile performs 2\,184\,000 inner steps over
44.98 hours, whereas STRAINER performs 5000 steps in approximately
0.66 hours. The numerical advantage of Reptile therefore cannot be
separated from the effect of its larger budget or interpreted as an
advantage under equal computational cost.

\subsection{Reptile Versus Random Initialization}

\begin{figure}[!htbp]
  \centering
  \DicePlot{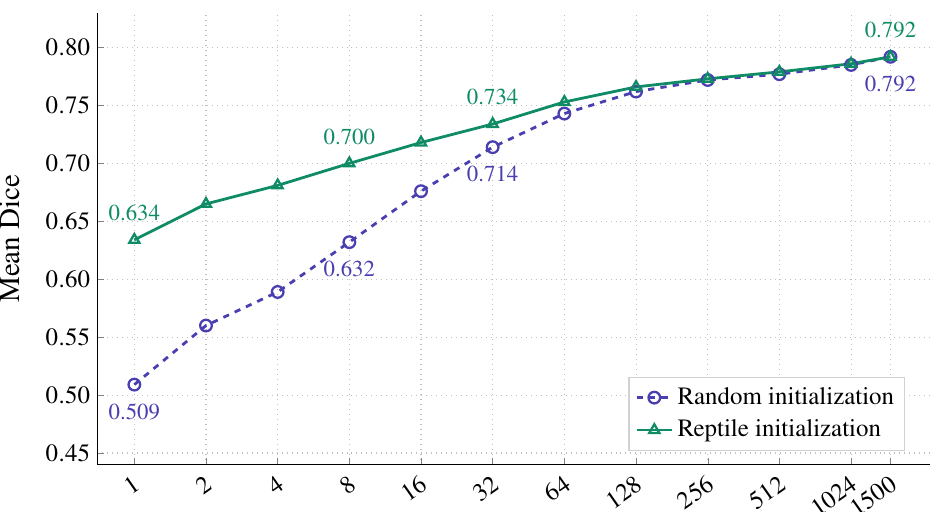}
  \HDPlot{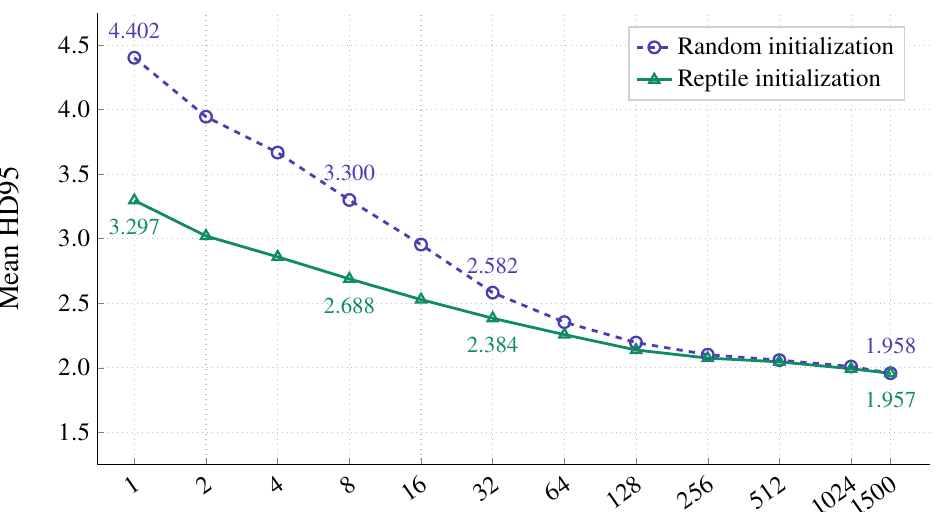}
  \caption{Mean Dice and HD95 versus the number of adaptation iterations
  for KAN-IDIR on OASIS-1 with random and Reptile initialization.}
  \label{fig:reptile_kan}
\end{figure}

\begin{figure}[!htbp]
  \centering
  \DicePlot{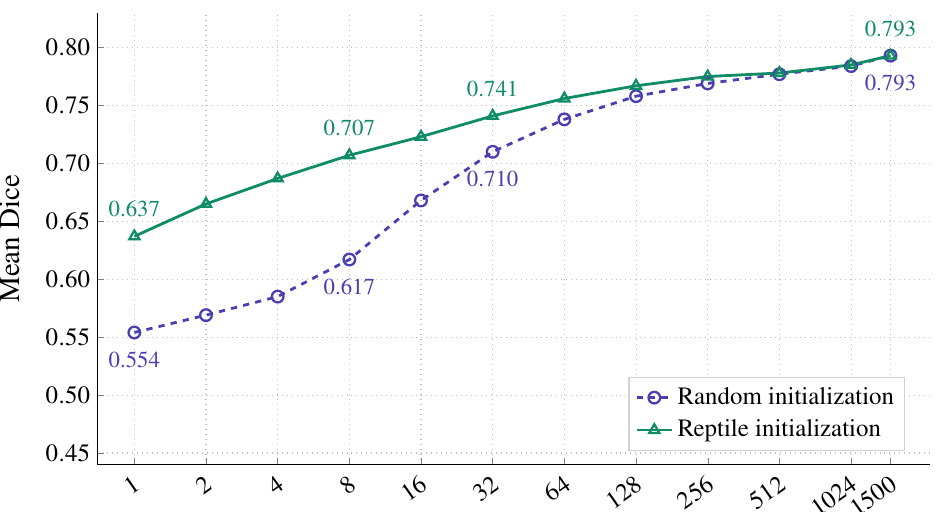}
  \HDPlot{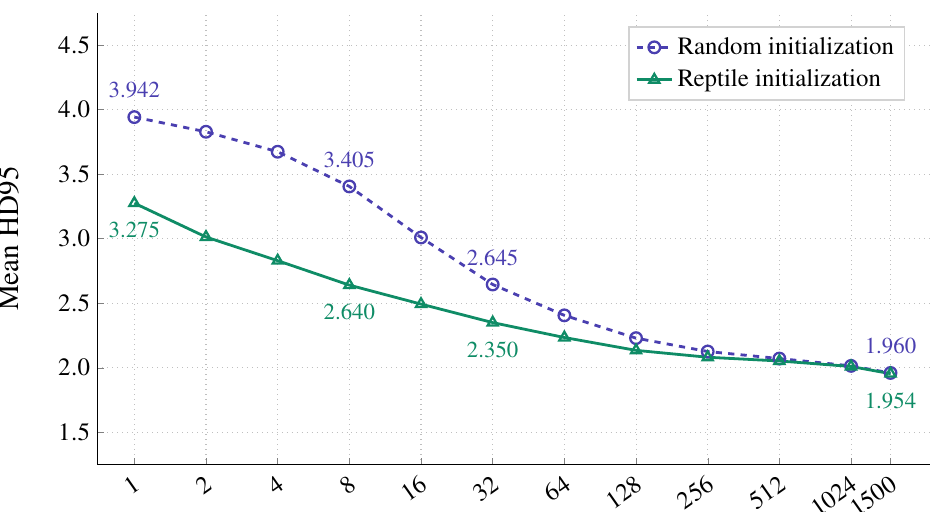}
  \caption{Mean Dice and HD95 versus the number of adaptation iterations
  for RandKAN-IDIR on OASIS-1 with random and Reptile initialization.}
  \label{fig:reptile_randkan}
\end{figure}

Figures~\ref{fig:reptile_kan} and~\ref{fig:reptile_randkan} show mean Dice
and HD95 as functions of the number of adaptation iterations for random
and Reptile initialization of the two base coordinate models. The main
effect of meta-initialization occurs early in optimization: for both
models, Reptile yields higher Dice and lower HD95 from the first
iterations. As the number of iterations increases, the gap gradually
narrows, and the curves for both metrics approach similar values by
the end of optimization. This indicates that meta-initialization
primarily accelerates adaptation and has little effect on the final
quality when optimization is sufficiently long.

\FloatBarrier
\subsection{STRAINER Versus Other Initialization Schemes}

\begin{figure}[!htbp]
  \centering
  \DicePlot{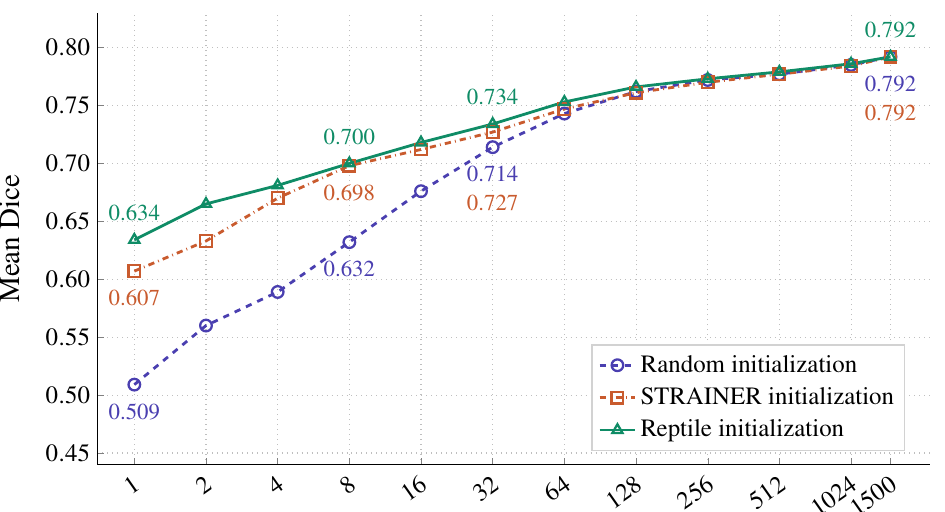}
  \HDPlot{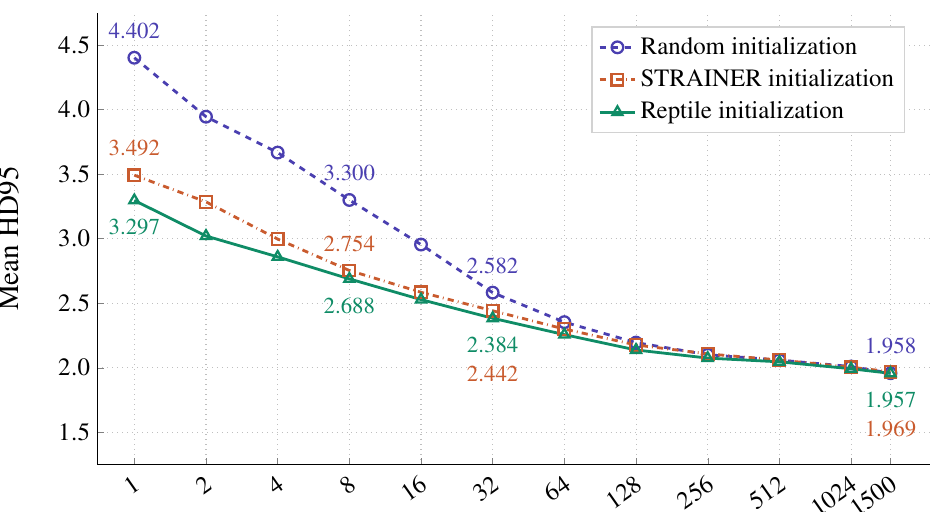}
  \caption{Mean Dice and HD95 versus the number of adaptation iterations
  for KAN-IDIR on OASIS-1 with random, STRAINER, and Reptile initialization.}
  \label{fig:strainer}
\end{figure}

STRAINER was evaluated separately. Its practical distinction is the low
pretraining cost: in our experiments, pretraining took approximately
39--40 minutes (about 0.66 hours), whereas Reptile meta-training required
approximately 2699 minutes (44.98 hours), making STRAINER pretraining
about 68 times faster. Figure~\ref{fig:strainer} compares the three
schemes. In the configuration used, STRAINER yields higher mean Dice
and lower mean HD95 than random initialization from the first
iterations and rapidly approaches Reptile initialization. With further
optimization, the differences among all three variants decrease.

\FloatBarrier
The fraction of voxels with a negative Jacobian determinant was additionally
evaluated at the final iteration, 1500 (Table~\ref{tab:folding}). For
KAN-IDIR, the lowest mean fraction was obtained with STRAINER
initialization. For RandKAN-IDIR, Reptile initialization also reduces
folding compared with random initialization.

\begin{table}[tbp]
  \centering
  \caption{Fraction of voxels with a negative Jacobian determinant at
  iteration 1500, reported as mean $\pm$ standard deviation (\%).}
  \label{tab:folding}
  \setlength{\tabcolsep}{12pt}
  \begin{tabular}{llc}
    \toprule
    Model & Initialization & Fraction (\%) \\
    \midrule
    KAN-IDIR & Random & $0.05005\pm0.02631$ \\
    KAN-IDIR & STRAINER & $0.02168\pm0.01519$ \\
    KAN-IDIR & Reptile & $0.02560\pm0.01454$ \\
    RandKAN-IDIR & Random & $0.04976\pm0.04542$ \\
    RandKAN-IDIR & Reptile & $0.03869\pm0.02063$ \\
    \bottomrule
  \end{tabular}
\end{table}

To quantify the adaptation-speed gain, Table~\ref{tab:speed} reports
Dice at selected iterations and the number of iterations required to
reach $\text{Dice}\geq0.70$ (values read from the curves in
Fig.~\ref{fig:strainer}). Meta-initialization reaches the target quality
in fewer iterations: on the evaluation grid used, the $0.70$ threshold
is reached at approximately iteration 8 with Reptile, iteration 16 with
STRAINER, and iteration 32 with random initialization. Because the
threshold was checked only at powers of two, the actual ratio of
iteration counts for random and Reptile initialization lies between 2
and 8; it has not been established to equal 4.

\begin{table}[tbp]
  \centering
  \caption{Dice and iterations required to reach the $\text{Dice}=0.70$
  threshold for different initialization schemes on OASIS-1.}
  \label{tab:speed}
  \setlength{\tabcolsep}{8pt}
  \begin{tabular}{lccccc}
    \toprule
    Initialization & Iter.\ 1 & Iter.\ 8 & Iter.\ 32 & Iter.\ 64 & Iter.\ to 0.70 \\
    \midrule
    Random & 0.509 & 0.632 & 0.714 & 0.743 & $\approx32$ \\
    STRAINER & 0.607 & 0.698 & 0.727 & 0.747 & $\approx16$ \\
    Reptile & 0.634 & 0.700 & 0.734 & 0.753 & $\approx8$ \\
    \bottomrule
  \end{tabular}
\end{table}

In the selected configuration, complete optimization of one pair
(1500 iterations) takes approximately 60 seconds, corresponding to about
$0.04$ seconds per adaptation iteration on average. This per-iteration
value is obtained by dividing the total time by the number of iterations
and excludes fixed overheads. Reaching $\text{Dice}=0.70$ therefore takes
an estimated $1.3$ seconds with random initialization and $0.3$ seconds
with Reptile initialization. Meta-initialization,
however, incurs a one-time pretraining cost that is amortized when
many image pairs are subsequently registered. STRAINER is considerably
less expensive than Reptile in this respect.

Based on the discrete checkpoints in the table, the estimated savings
relative to random initialization are approximately 0.96 seconds per
pair for Reptile and 0.64 seconds for STRAINER. Accounting for pretraining
cost gives an estimated break-even point of approximately 170\,000
pairs for Reptile and 3700 pairs for STRAINER. For comparison, complete
optimization of 413 pairs at 60 seconds per pair takes approximately
6.9 hours. Reptile pretraining costs approximately 6.5 times that amount,
whereas STRAINER pretraining costs about 10\,\% of it. According to the
reported curves, both schemes require
approximately 1024 iterations to reach a target Dice of $0.785$, so the
early-adaptation advantage has largely disappeared at this quality level.

\FloatBarrier
\subsection{Visual Comparison After a Small Number of Iterations}

\begin{figure}[H]
  \centering
  \includegraphics[width=0.85\linewidth]{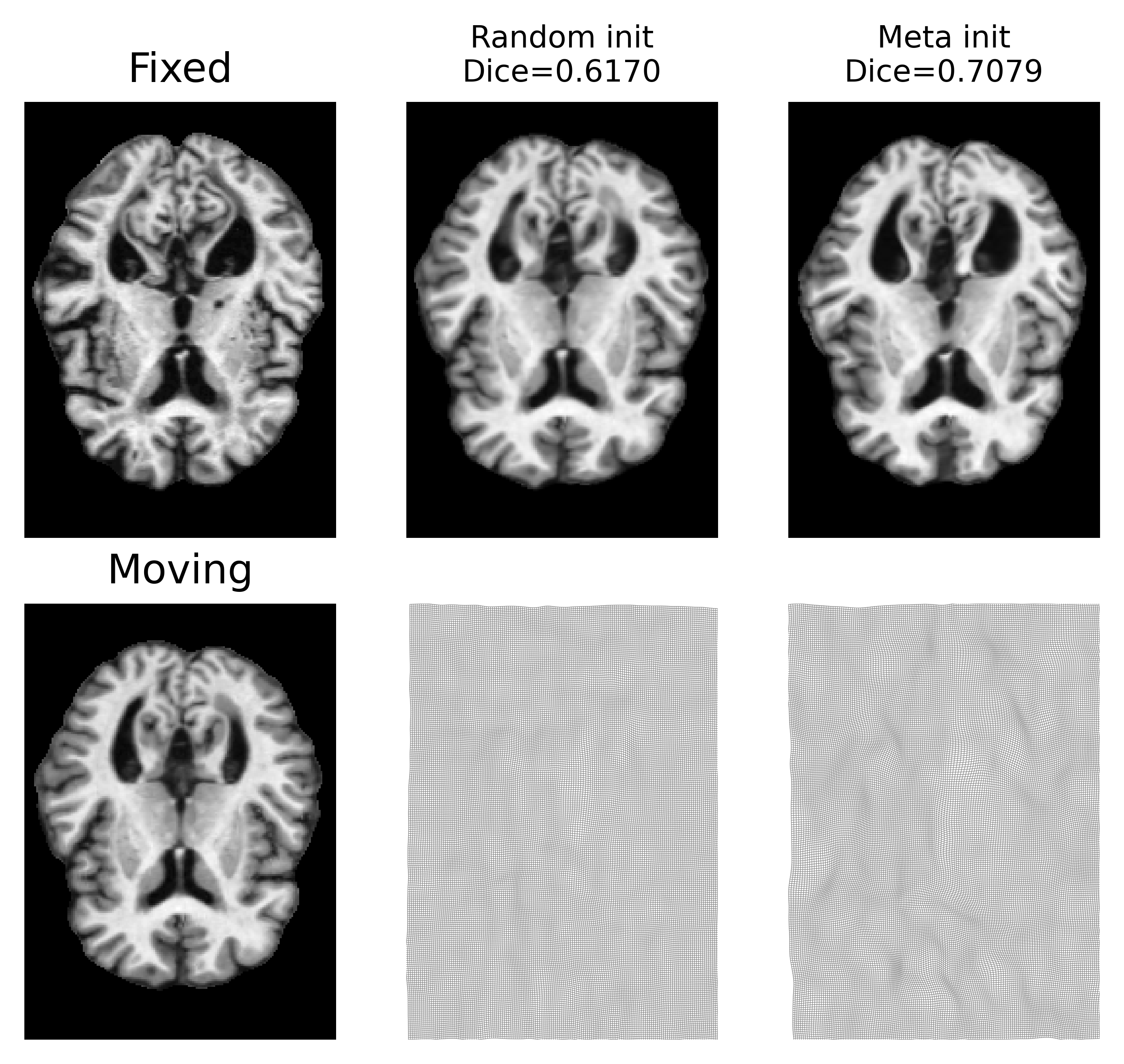}
  \caption{Visual comparison of registration after 10 adaptation
  iterations with random and Reptile meta-initialization on OASIS-1.
  Top row: the fixed image and the two warped images. Bottom row: the
  moving image and deformation fields for the two initializations.}
  \label{fig:visual}
\end{figure}

The visual example in Fig.~\ref{fig:visual} is consistent with the
quantitative results. With the same small number of iterations (10),
Reptile meta-initialization produces a closer match to the fixed image
than random initialization (Dice $0.7079$ versus $0.6170$). This single
example illustrates that the benefit of meta-initialization is most
pronounced under a limited iteration budget.

\section{Discussion}

The results support the following conclusions. First, INR meta-initialization
is primarily useful when the adaptation-iteration budget is limited:
with few iterations, Reptile and STRAINER yield higher mean Dice and
lower mean HD95 than random initialization. Second, when the full
iteration budget is used, this advantage largely disappears. All three
schemes converge to similar values for both metrics, so the effect of
meta-learning should be interpreted as faster adaptation rather than
an improvement in the final registration quality.

Third, the two schemes differ in early-iteration quality and pretraining
cost. Reptile achieves somewhat higher Dice at the beginning of adaptation,
whereas STRAINER pretraining was approximately 68 times faster in the
experiments and STRAINER also outperformed random initialization in mean
Dice. However, the unmatched budgets and the absence of variability
estimates prevent these results from establishing which method is better
under equal computational cost. Both schemes rely on similarity between
the distributions of training and test pairs: under a strong distribution
shift, parameter transferability and the resulting speed gain may
decrease. This dependence was not tested in the present experiment.

LUMINR \cite{rynczak2026luminr} also reports accelerated adaptation with
Reptile, but uses a locally conditioned SIREN rather than a coordinate-only
KAN. It is also evaluated on its own random split of OASIS into 280
training, 34 validation, and 100 test volumes. Absolute Dice
values and speedups are not directly comparable because
the architectures, data splits, label conventions, and optimization settings
differ. A controlled comparison under a common evaluation protocol would be
needed to assess their relative performance.

\section{Conclusion}

In the selected configurations, Reptile and STRAINER yield higher mean
Dice and lower mean HD95 at early iterations than random initialization,
while the values approach one another with longer adaptation. STRAINER
pretraining requires substantially less time, but the unmatched budgets
and the absence of variability estimates for the main metrics preclude
ranking the methods under equal computational cost.

Future work includes comparisons at matched budgets, evaluation of
SDlogJ and deformation-field regularity at early iterations, analysis
of variability across pairs and random seeds, additional baselines,
and assessment of transferability to other distributions and datasets.

\section*{Acknowledgments}
This research was supported by the Russian Science Foundation,
grant No.~26-11-00118, \url{https://rscf.ru/project/26-11-00118/}.

\bibliographystyle{unsrtnat}
\bibliography{references}

\begin{thebibliography}{44}
\providecommand{\natexlab}[1]{#1}
\providecommand{\url}[1]{\texttt{#1}}
\expandafter\ifx\csname urlstyle\endcsname\relax
  \providecommand{\doi}[1]{doi: #1}\else
  \providecommand{\doi}{doi: \begingroup \urlstyle{rm}\Url}\fi

\bibitem[Rueckert et~al.(1999)]{Rueckert1999}
Daniel Rueckert et~al.
\newblock {Nonrigid Registration Using Free-Form Deformations: Application to
  Breast MR Images}.
\newblock \emph{IEEE Transactions on Medical Imaging}, 18\penalty0
  (8):\penalty0 712--721, 1999.

\bibitem[Beg et~al.(2005)]{Beg2005}
Mirza~Faisal Beg et~al.
\newblock {Computing Large Deformation Metric Mappings via Geodesic Flows of
  Diffeomorphisms}.
\newblock \emph{International Journal of Computer Vision}, 61\penalty0
  (2):\penalty0 139--157, 2005.

\bibitem[Thirion(1998)]{Thirion1998}
Jean-Philippe Thirion.
\newblock {Image Matching as a Diffusion Process: An Analogy with Maxwell's
  Demons}.
\newblock \emph{Medical Image Analysis}, 2\penalty0 (3):\penalty0 243--260,
  1998.

\bibitem[Klein et~al.(2010)]{Klein2009}
Stefan Klein et~al.
\newblock {Elastix: A Toolbox for Intensity-Based Medical Image Registration}.
\newblock \emph{IEEE Transactions on Medical Imaging}, 29\penalty0
  (1):\penalty0 196--205, 2010.

\bibitem[Avants et~al.(2011)]{Avants2011}
Brian~B. Avants et~al.
\newblock {A Reproducible Evaluation of {ANTs} Similarity Metric Performance in
  Brain Image Registration}.
\newblock \emph{NeuroImage}, 54\penalty0 (3):\penalty0 2033--2044, 2011.

\bibitem[Vishnevskiy et~al.(2017)]{Vishnevskiy2016}
Valery Vishnevskiy et~al.
\newblock {Isotropic Total Variation Regularization of Displacements in
  Parametric Image Registration}.
\newblock \emph{IEEE Transactions on Medical Imaging}, 36\penalty0
  (2):\penalty0 385--395, 2017.

\bibitem[Zhang and Fletcher(2019)]{Zhang2019}
Miaomiao Zhang and P.~Thomas Fletcher.
\newblock {Fast Diffeomorphic Image Registration via Fourier-Approximated Lie
  Algebras}.
\newblock \emph{International Journal of Computer Vision}, 127\penalty0
  (1):\penalty0 61--73, 2019.

\bibitem[Balakrishnan et~al.(2019)]{Balakrishnan2019a}
Guha Balakrishnan et~al.
\newblock {{VoxelMorph}: A Learning Framework for Deformable Medical Image
  Registration}.
\newblock \emph{IEEE Transactions on Medical Imaging}, 38\penalty0
  (8):\penalty0 1788--1800, 2019.

\bibitem[Chen et~al.(2022)]{Chen2022}
Junyu Chen et~al.
\newblock {{TransMorph}: Transformer for Unsupervised Medical Image
  Registration}.
\newblock \emph{Medical Image Analysis}, 82:\penalty0 102615, 2022.

\bibitem[Mildenhall et~al.(2022)]{Mildenhall2020}
Ben Mildenhall et~al.
\newblock {{NeRF}: Representing Scenes as Neural Radiance Fields for View
  Synthesis}.
\newblock \emph{Communications of the ACM}, 65\penalty0 (1):\penalty0 99--106,
  2022.

\bibitem[Wolterink et~al.(2022)]{Wolterink2022}
Jelmer~M. Wolterink et~al.
\newblock {Implicit Neural Representations for Deformable Image Registration}.
\newblock In \emph{{International Conference on Medical Imaging with Deep
  Learning}}, pages 1349--1359. PMLR, 2022.

\bibitem[van Harten et~al.(2024{\natexlab{a}})]{Harten2023}
Louis~D. van Harten et~al.
\newblock {Robust Deformable Image Registration Using Cycle-Consistent Implicit
  Representations}.
\newblock \emph{IEEE Transactions on Medical Imaging}, 43\penalty0
  (2):\penalty0 784--793, 2024{\natexlab{a}}.

\bibitem[van Harten et~al.(2024{\natexlab{b}})]{Harten2024}
Louis van Harten et~al.
\newblock {Deformable Image Registration with Geometry-Informed Implicit Neural
  Representations}.
\newblock In \emph{{Medical Imaging with Deep Learning}}, pages 730--742. PMLR,
  2024{\natexlab{b}}.

\bibitem[Nichol et~al.(2018)Nichol, Achiam, and Schulman]{Nichol2018}
Alex Nichol, Joshua Achiam, and John Schulman.
\newblock {On First-Order Meta-Learning Algorithms}, 2018.
\newblock URL \url{https://arxiv.org/abs/1803.02999}.
\newblock arXiv:1803.02999.

\bibitem[Vyas et~al.(2024)]{Vyas2024Strainer}
Kushal Vyas et~al.
\newblock {Learning Transferable Features for Implicit Neural Representations}.
\newblock \emph{Advances in Neural Information Processing Systems},
  37:\penalty0 42268--42291, 2024.

\bibitem[Dalca et~al.(2019)]{Dalca2019}
Adrian~V. Dalca et~al.
\newblock {Unsupervised Learning of Probabilistic Diffeomorphic Registration
  for Images and Surfaces}.
\newblock \emph{Medical Image Analysis}, 57:\penalty0 226--236, 2019.

\bibitem[Mok and Chung(2020{\natexlab{a}})]{Mok2020b}
Tony C.~W. Mok and Albert C.~S. Chung.
\newblock {Large Deformation Diffeomorphic Image Registration with Laplacian
  Pyramid Networks}.
\newblock In \emph{{International Conference on Medical Image Computing and
  Computer-Assisted Intervention}}, pages 211--221, Cham, 2020{\natexlab{a}}.
  Springer.

\bibitem[Mok and Chung(2020{\natexlab{b}})]{Mok2020a}
Tony C.~W. Mok and Albert C.~S. Chung.
\newblock {Fast Symmetric Diffeomorphic Image Registration with Convolutional
  Neural Networks}.
\newblock In \emph{{Proceedings of the IEEE/CVF Conference on Computer Vision
  and Pattern Recognition}}, pages 4644--4653, 2020{\natexlab{b}}.

\bibitem[Greer et~al.(2021)]{Greer2021}
Hastings Greer et~al.
\newblock {{ICON}: Learning Regular Maps through Inverse Consistency}.
\newblock In \emph{{Proceedings of the IEEE/CVF International Conference on
  Computer Vision}}, pages 3396--3405, 2021.

\bibitem[Tian et~al.(2023)]{Tian2023}
Lin Tian et~al.
\newblock {{GradICON}: Approximate Diffeomorphisms via Gradient Inverse
  Consistency}.
\newblock In \emph{{Proceedings of the IEEE/CVF Conference on Computer Vision
  and Pattern Recognition}}, pages 18084--18094, 2023.

\bibitem[Meng et~al.(2024)]{Meng2024}
Mingyuan Meng et~al.
\newblock {Correlation-Aware Coarse-to-Fine {MLP}s for Deformable Medical Image
  Registration}.
\newblock In \emph{{Proceedings of the IEEE/CVF Conference on Computer Vision
  and Pattern Recognition}}, pages 9645--9654, 2024.

\bibitem[Jena et~al.(2024)]{Jena2024}
Rohit Jena et~al.
\newblock {Deep Learning in Medical Image Registration: Magic or Mirage?}
\newblock \emph{Advances in Neural Information Processing Systems},
  37:\penalty0 108331--108353, 2024.

\bibitem[Tian et~al.(2024)]{Tian2024}
Lin Tian et~al.
\newblock {{uniGradICON}: A Foundation Model for Medical Image Registration}.
\newblock In \emph{{International Conference on Medical Image Computing and
  Computer-Assisted Intervention}}, pages 749--760, Cham, 2024. Springer.

\bibitem[Drozdov and Sorokin(2024)]{Drozdov2024}
Nikita~A. Drozdov and Dmitry~V. Sorokin.
\newblock {{FNOReg}: Resolution-Robust Medical Image Registration Method Based
  on Fourier Neural Operator}.
\newblock In \emph{{International Conference on Pattern Recognition}}, pages
  163--177, Cham, 2024. Springer.

\bibitem[Byra et~al.(2023)]{Byra2023}
Micha{\l} Byra et~al.
\newblock {Exploring the Performance of Implicit Neural Representations for
  Brain Image Registration}.
\newblock \emph{Scientific Reports}, 13\penalty0 (1):\penalty0 17334, 2023.

\bibitem[Wu et~al.(2022)]{Li2021}
Yifan Wu et~al.
\newblock {{NODEO}: A Neural Ordinary Differential Equation Based Optimization
  Framework for Deformable Image Registration}.
\newblock In \emph{{Proceedings of the IEEE/CVF Conference on Computer Vision
  and Pattern Recognition}}, pages 20804--20813, 2022.

\bibitem[Sideri-Lampretsa et~al.(2024)]{SideriLampretsa2024}
Vasiliki Sideri-Lampretsa et~al.
\newblock {{SINR}: Spline-Enhanced Implicit Neural Representation for
  Multi-Modal Registration}.
\newblock In \emph{{Medical Imaging with Deep Learning}}, pages 1462--1474,
  2024.

\bibitem[Finn et~al.(2017)]{Finn2017}
Chelsea Finn et~al.
\newblock {Model-Agnostic Meta-Learning for Fast Adaptation of Deep Networks}.
\newblock In \emph{{International Conference on Machine Learning}}, pages
  1126--1135. PMLR, 2017.

\bibitem[Li et~al.(2017)]{Li2017MetaSGD}
Zhenguo Li et~al.
\newblock {{Meta-SGD}: Learning to Learn Quickly for Few-Shot Learning}, 2017.
\newblock URL \url{https://arxiv.org/abs/1707.09835}.
\newblock arXiv:1707.09835.

\bibitem[Sitzmann et~al.(2020)]{MetaSDF2020}
Vincent Sitzmann et~al.
\newblock {{MetaSDF}: Meta-Learning Signed Distance Functions}.
\newblock \emph{Advances in Neural Information Processing Systems},
  33:\penalty0 10136--10147, 2020.

\bibitem[Tancik et~al.(2021)]{Tancik2021}
Matthew Tancik et~al.
\newblock {Learned Initializations for Optimizing Coordinate-Based Neural
  Representations}.
\newblock In \emph{{Proceedings of the IEEE/CVF Conference on Computer Vision
  and Pattern Recognition}}, pages 2846--2855, 2021.

\bibitem[Lee et~al.(2021)]{Lee2021MetaSparseINR}
Jaeho Lee et~al.
\newblock {Meta-Learning Sparse Implicit Neural Representations}.
\newblock \emph{Advances in Neural Information Processing Systems},
  34:\penalty0 11769--11780, 2021.

\bibitem[Str{\"u}mpler et~al.(2022)]{Strumpler2022}
Yannick Str{\"u}mpler et~al.
\newblock {Implicit Neural Representations for Image Compression}.
\newblock In \emph{{European Conference on Computer Vision}}, pages 74--91,
  Cham, 2022. Springer.

\bibitem[van Harten et~al.(2024{\natexlab{c}})]{Harten2024Reindir}
Louis van Harten et~al.
\newblock {{REINDIR}: Repeated Embedding Infusion for Neural Deformable Image
  Registration}.
\newblock In \emph{{Medical Imaging with Deep Learning}}, pages 577--595. PMLR,
  2024{\natexlab{c}}.

\bibitem[Qian et~al.(2026)]{Qian2026}
Xiaoxue Qian et~al.
\newblock {{cMeta-INR}: Cohort-Informed Meta-Learning-Based Implicit Neural
  Representation for Deformable Registration-Driven Real-Time Volumetric {MRI}
  Estimation}.
\newblock \emph{Physics in Medicine \& Biology}, 71\penalty0 (1):\penalty0
  015003, 2026.

\bibitem[Rynczak and Wodzinski(2026)]{rynczak2026luminr}
Damian Rynczak and Marek Wodzinski.
\newblock {{LUMINR}: Intensity-Conditioned Meta-Learned {INR}s for Accelerated
  Medical Image Registration}, 2026.
\newblock URL \url{https://arxiv.org/abs/2610.24046}.
\newblock arXiv:2610.24046.

\bibitem[Drozdov et~al.(2026)Drozdov, Zinovev, and Sorokin]{make8070184}
Nikita~A. Drozdov, Marat~O. Zinovev, and Dmitry~V. Sorokin.
\newblock {Deformable Medical Image Registration with {KAN}-Based Implicit
  Neural Representations}.
\newblock \emph{Machine Learning and Knowledge Extraction}, 8\penalty0
  (7):\penalty0 184, 2026.
\newblock \doi{10.3390/make8070184}.

\bibitem[Liu et~al.(2024)]{Liu2024c}
Ziming Liu et~al.
\newblock {{KAN}: Kolmogorov--Arnold Networks}, 2024.
\newblock URL \url{https://arxiv.org/abs/2404.19756}.
\newblock arXiv:2404.19756.

\bibitem[SS et~al.(2024)]{Sidharth2024}
Sidharth SS et~al.
\newblock {Chebyshev Polynomial-Based Kolmogorov--Arnold Networks: An Efficient
  Architecture for Nonlinear Function Approximation}, 2024.
\newblock URL \url{https://arxiv.org/abs/2405.07200}.
\newblock arXiv:2405.07200.

\bibitem[Rudin et~al.(1992)Rudin, Osher, and Fatemi]{Rudin1992}
Leonid~I. Rudin, Stanley Osher, and Emad Fatemi.
\newblock {Nonlinear Total Variation Based Noise Removal Algorithms}.
\newblock \emph{Physica D: Nonlinear Phenomena}, 60\penalty0 (1--4):\penalty0
  259--268, 1992.

\bibitem[Marcus et~al.(2007)]{Marcus2007}
Daniel~S. Marcus et~al.
\newblock {Open Access Series of Imaging Studies ({OASIS}): Cross-Sectional
  {MRI} Data in Young, Middle Aged, Nondemented, and Demented Older Adults}.
\newblock \emph{Journal of Cognitive Neuroscience}, 19\penalty0 (9):\penalty0
  1498--1507, 2007.

\bibitem[Hoopes et~al.(2022)]{Hoopes2022}
Andrew Hoopes et~al.
\newblock {Learning the Effect of Registration Hyperparameters with
  {HyperMorph}}.
\newblock \emph{Machine Learning for Biomedical Imaging}, 1:\penalty0 1--30,
  2022.

\bibitem[Paszke et~al.(2019)]{Paszke2019}
Adam Paszke et~al.
\newblock {{PyTorch}: An Imperative Style, High-Performance Deep Learning
  Library}.
\newblock \emph{Advances in Neural Information Processing Systems}, 32, 2019.

\bibitem[Kingma and Ba(2014)]{Kingma2014}
Diederik~P. Kingma and Jimmy Ba.
\newblock {Adam: A Method for Stochastic Optimization}, 2014.
\newblock URL \url{https://arxiv.org/abs/1412.6980}.
\newblock arXiv:1412.6980.

\end{thebibliography}
\end{document}